\documentclass[runningheads]{llncs}
\usepackage{graphicx}

\usepackage{tikz}
\usepackage{comment}
\usepackage{amsmath,amssymb} 
\usepackage{color}
\usepackage{booktabs}
\usepackage{hyperref}

\begin{document}
\pagestyle{headings}
\mainmatter
\def\ECCVSubNumber{35}  

\title{AB3DMOT: A Baseline for 3D Multi-Object Tracking and New Evaluation Metrics} 

\titlerunning{AB3DMOT: A Baseline for 3D Multi-Object Tracking}
%
\author{Xinshuo Weng \and
Jianren Wang \and
David Held \and
Kris Kitani}
\authorrunning{X. Weng et al.}
%
\institute{Robotics Institute, Carnegie Mellon University, Pittsburgh, USA\\
\email{\{xinshuow, jianrenw, dheld, kkitani\}@cs.cmu.edu}}
\maketitle

\begin{abstract}
\vspace{-0.6cm}
3D multi-object tracking (MOT) is essential to applications such as autonomous driving. Recent work focuses on developing accurate systems giving less attention to computational cost and system complexity. In contrast, this work proposes a simple real-time 3D MOT system with strong performance. Our system first obtains 3D detections from a LiDAR point cloud. Then, a straightforward combination of a 3D Kalman filter and the Hungarian algorithm is used for state estimation and data association. Additionally, 3D MOT datasets such as KITTI evaluate MOT methods in 2D space and standardized 3D MOT evaluation tools are missing for a fair comparison of 3D MOT methods. We propose a new 3D MOT evaluation tool along with three new metrics to comprehensively evaluate 3D MOT methods. We show that, our proposed method achieves strong 3D MOT performance on KITTI and runs at a rate of $207.4$ FPS on the KITTI dataset, achieving the fastest speed among modern 3D MOT systems. Our code is publicly available at \url{http://www.xinshuoweng.com/projects/AB3DMOT}.
\vspace{-0.2cm}
\vspace{-0.4cm}
\end{abstract}

\vspace{-0.7cm}
\section{Introduction}
\vspace{-0.25cm}

\begin{figure}[t]
\centering
\includegraphics[trim=0.3cm 2.5cm 9.5cm 0.3cm, clip=true, width=0.345\linewidth]{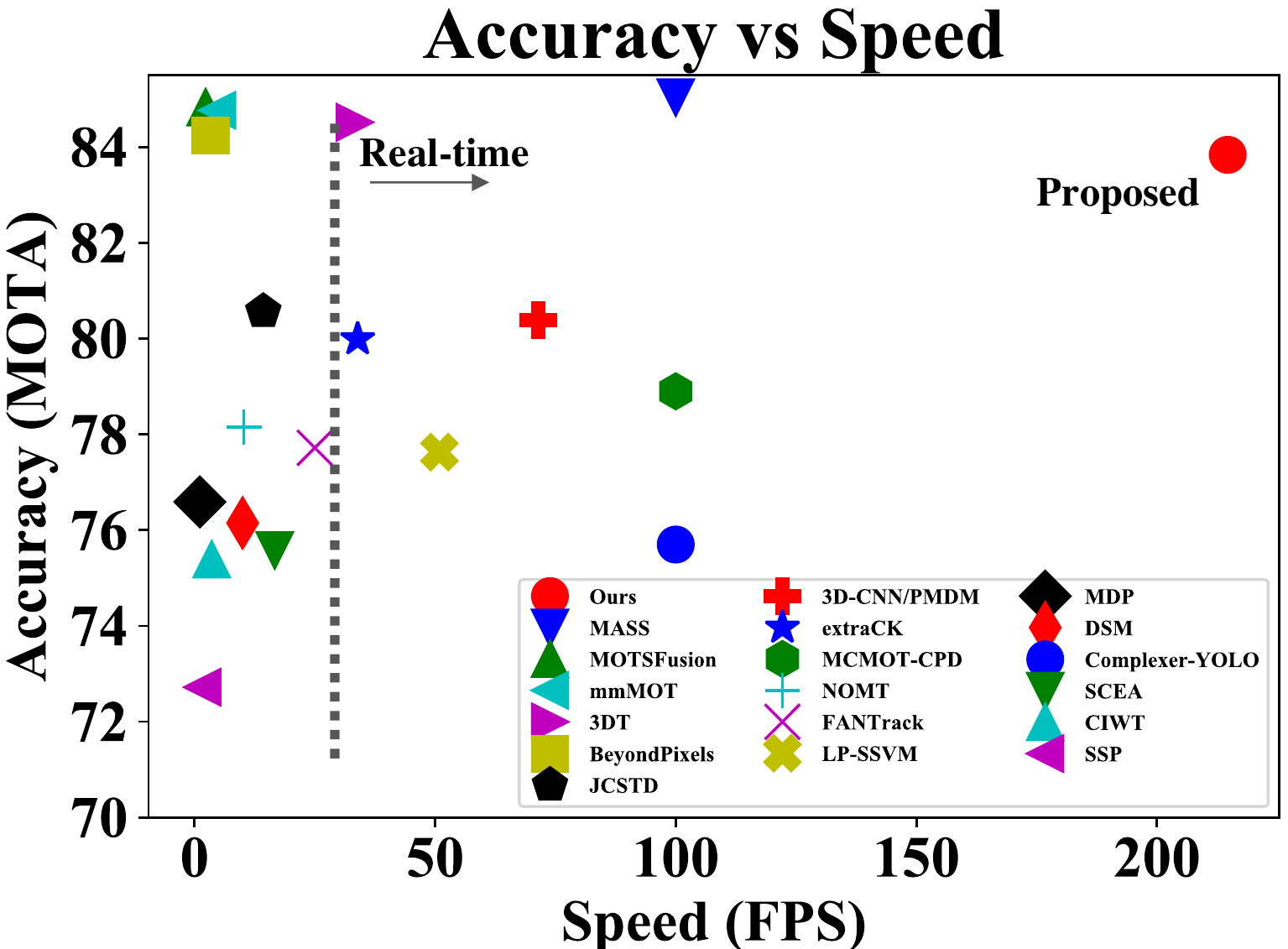}
\includegraphics[trim=0cm 7cm 5.2cm 0.1cm, clip=true, width=0.645\linewidth]{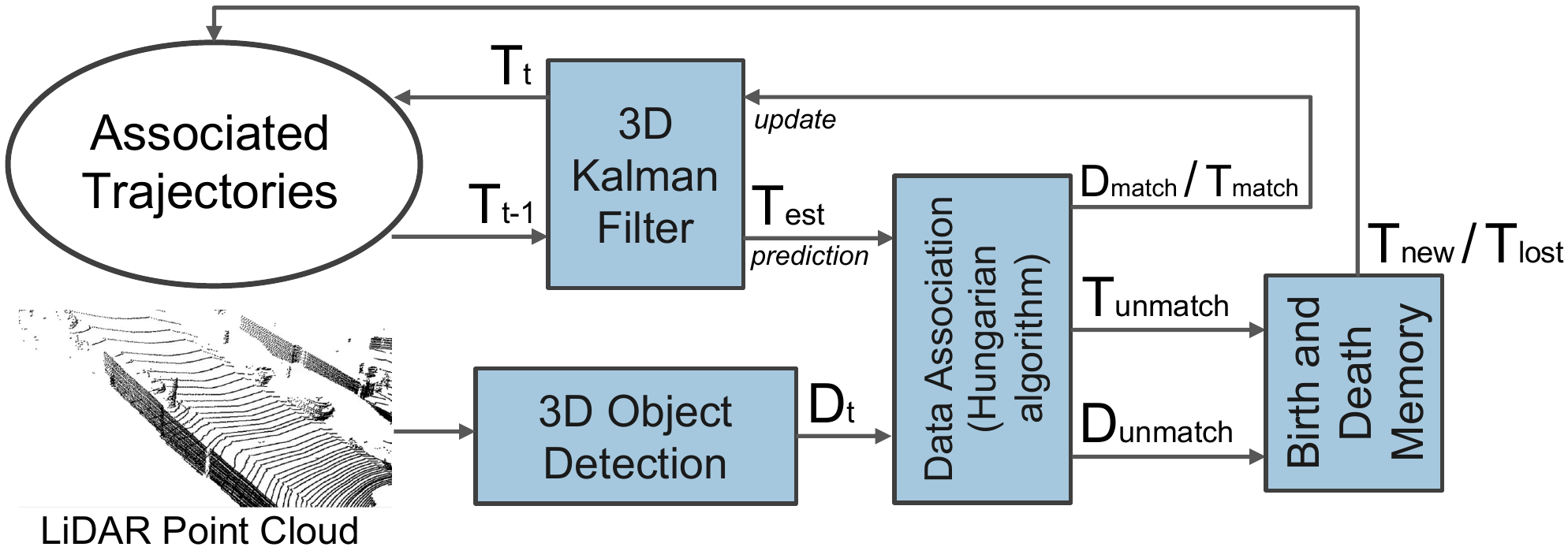}
\vspace{-0.8cm}
\caption{\textbf{Left}: MOTA of modern MOT systems on KITTI 2D MOT leaderboard. The higher and more right is better. \textbf{Right}: Proposed system pipeline.}
\vspace{-0.3cm}
\label{fig:teaser}
\end{figure}

MOT is essential to applications such as autonomous driving \cite{Wang2018}. Due to advancements in detection, there has been much progress on MOT. For example, for the car class on the KITTI~\cite{Geiger2012} 2D MOT benchmark as shown in Fig. \ref{fig:teaser} (Left), the MOTA (multi-object tracking accuracy) has improved from 57.03 to 84.04 in just two years. While we are encouraged by the progress, we observe that our focus on innovation and accuracy may have come at the cost of important practical factors such as computational efficiency and system simplicity, and S.O.T.A. methods typically require a large computational cost \cite{Weng2020_gnntrkforecast,Baser2019,Weng2020_gnn3dmot} making real-time performance a challenge. Also, modern MOT systems are often complex and it is not always clear which part of the system contributes the most to performance. 

\vspace{-0.5cm}
\section{AB3DMOT: A Baseline for 3D MOT}
\vspace{-0.25cm}

To provide a standard 3D MOT baseline for comparative analysis, we implement a classical approach which is both efficient and simple in design -- the Kalman filter \cite{Kalman1960} (1960) coupled with the Hungarian method \cite{WKuhn1955} (1955). Specifically, our system is shown in Fig. \ref{fig:teaser} (Right), which employs an off-the-shelf 3D object detector to obtain 3D detections from the LiDAR point cloud \cite{Shi2019}. Then, a combination of the 3D Kalman filter (with a constant velocity model) and the Hungarian algorithm is used for state estimation and data association. Unlike other filter-based MOT systems which define the state space of the filter in the image plane \cite{Wojke2017}, we extend the state space of the objects to 3D, including 3D location, 3D size, 3D velocity and orientation. 

Our empirical results are alarming. While the combination of modules in our system is straightforward, we achieve strong 3D MOT performance on the KITTI dataset. Surprisingly, although our system does not use any 2D data as input, we also achieve competitive performance on the KITTI 2D MOT leaderboard in Fig. \ref{fig:teaser} (Left) by projecting our 3D MOT results onto the image plane for evaluation. 
We hypothesize that the strong 2D MOT performance of our 3D MOT system may be due to the fact that tracking in 3D can better resolve depth ambiguities and lead to fewer mismatches than tracking in 2D. Also, due to efficient design of our system, it runs at a rate of $207{.}4$ FPS on the KITTI dataset, achieving the fastest speed among modern 3D MOT systems. To be clear, the contribution of this work is not to innovate 3D MOT algorithms but to provide a more clear picture of modern 3D MOT systems in comparison to a most basic yet strong baseline, we believe the results of which are important to share across the community.

\vspace{-0.5cm}
\section{New 3D MOT Evaluation Tool}
\vspace{-0.25cm}

We observed one issue for current 3D MOT evaluation: \textit{Standard MOT benchmarks such as the KITTI dataset only support 2D MOT evaluation}, \emph{i.e.}, evaluation on the image plane. A tool to evaluate 3D MOT systems directly in 3D space is not currently available. On the KITTI dataset, the current convention of evaluating 3D MOT methods is to project the 3D tracking results to the 2D image plane and then use the KITTI 2D MOT evaluation tool. However, we believe that this will hamper the future progress of 3D MOT systems as evaluating on the image plane cannot demonstrate full strength of 3D MOT methods. 

To better evaluate 3D MOT systems, we implement an extension to the KITTI 2D MOT evaluation tool for 3D MOT evaluation. Specifically, we modify the cost function from 2D IoU to 3D IoU and match the 3D MOT results with 3D ground truth trajectories directly in the 3D space. In this way, we no longer need to project 3D MOT results to the image plane for evaluation. For every tracked object, a minimum 3D IoU (we use $0.25$ in our experiments) with the ground truth is required to be considered as a successful match. Although our 3D MOT evaluation tool is a straightforward extension to is 2D counterpart, we hope that it can serve as a standard to evaluate future 3D MOT systems.

\vspace{-0.5cm}
\section{New MOT Evaluation Metrics}
\vspace{-0.25cm}

Another issue we observed is that: \textit{Common MOT metrics such as MOTA and MOTP do not consider the confidence of tracked objects.} As a result, users must manually select a confidence threshold and filter out tracked objects with confidence lower than the threshold before evaluation. However, selecting the best threshold is non-trivial and the confidence threshold can be significantly different if using a different detector or evaluated on a different dataset. More importantly, using a single confidence threshold for evaluation prevents us from understanding the full spectrum of accuracy of a MOT system. One consequence is that a MOT system achieving high MOTA at a single threshold can still have extremely low MOTA at other thresholds, but still be ranked high on the leaderboard. Ideally, we should aim to develop MOT systems that can achieve high MOTA across a large set of thresholds, i.e., robust to the confidence score.

To deal with the issue that current MOT evaluation metrics do not consider the confidence score and only evaluate at a single threshold, we propose three integral metrics -- sAMOTA, AMOTA and AMOTP (scaled average MOTA, average MOTA and MOTP) -- to summarize the performance of MOTA and MOTP across different thresholds. Specifically, the integral metrics AMOTA and AMOTP are computed by integrating the MOTA and MOTP over all recall values. Similar to other integral metrics such as the average precision used in object detection, we approximate the integration with a summation over a discrete set (40) of recall values. Then, the sAMOTA matrix is proposed to adjust the range of the AMOTA value between $0\%$ and $100\%$.


\vspace{-0.5cm}
\section{Experiments}
\vspace{-0.25cm}

\begin{table*}[t]
\caption{Performance on KITTI val set using the proposed 3D MOT evaluation tool.}
\vspace{-0.3cm}
\centering
\resizebox{\textwidth}{!}{
\begin{tabular}{@{}llrrrrrrrrr@{}}
\toprule
Method \ \ \ \ \ \ \ \ \ \ & Input Data \ \ & \textbf{sAMOTA}$\uparrow$ & \ \ AMOTA$\uparrow$ & \ \ AMOTP$\uparrow$ & \ \ MOTA$\uparrow$ & \ \ MOTP$\uparrow$ & \ \ IDS$\downarrow$ & \ \ FRAG$\downarrow$ & \ \ FPS$\uparrow$ \\
\midrule
mmMOT~\cite{Zhang2019}      & 2D + 3D & 70.61 & 33.08 & 72.45 & 74.07 & 78.16 & 10  & 125 & 4.8 \\ 
FANTrack~\cite{Baser2019}   & 2D + 3D & 82.97 & 40.03 & 75.01 & 74.30 & 75.24 & 35  & 202 & 25.0 \\
\midrule
\textbf{Ours} & 3D & \textbf{93.28} & \textbf{45.43} & \textbf{77.41} & \textbf{86.24} & \textbf{78.43} & \textbf{0} & \textbf{15} & \textbf{207.4} \\
\bottomrule
\end{tabular}}
\vspace{-0.35cm}
\label{tab:3dcomparison}
\end{table*}

\begin{figure*}[t]
\begin{center}
\includegraphics[trim=0cm 6.6cm 0.1cm 0cm, clip=true, width=0.32\linewidth]{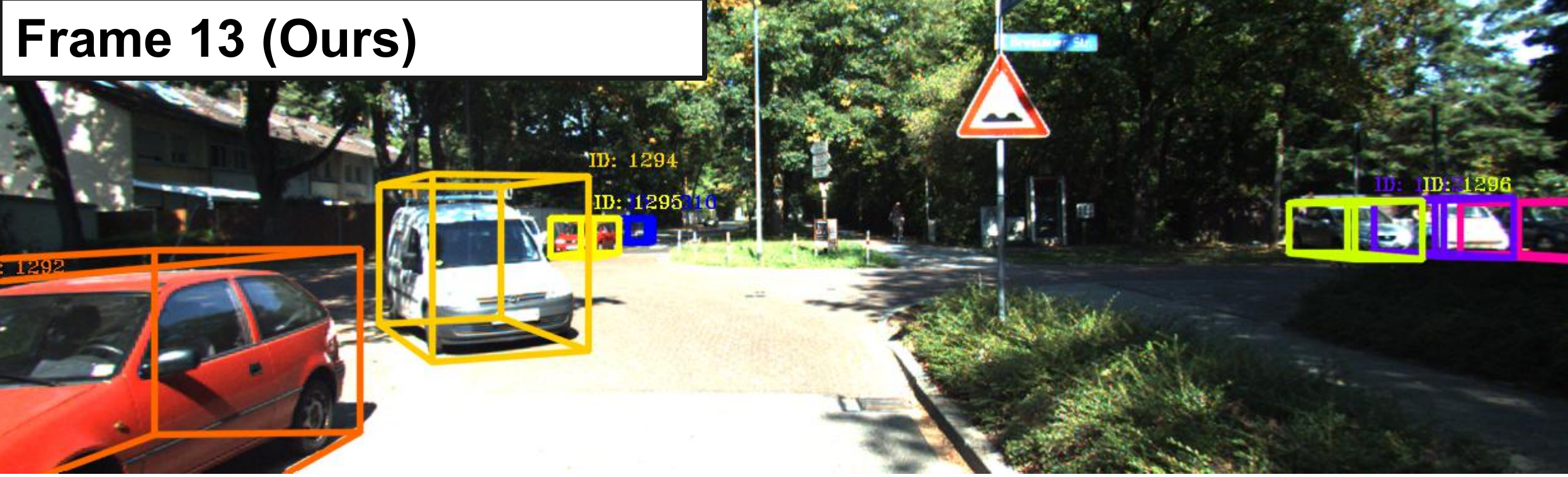}
\includegraphics[trim=0cm 6.6cm 0.1cm 0cm, clip=true, width=0.32\linewidth]{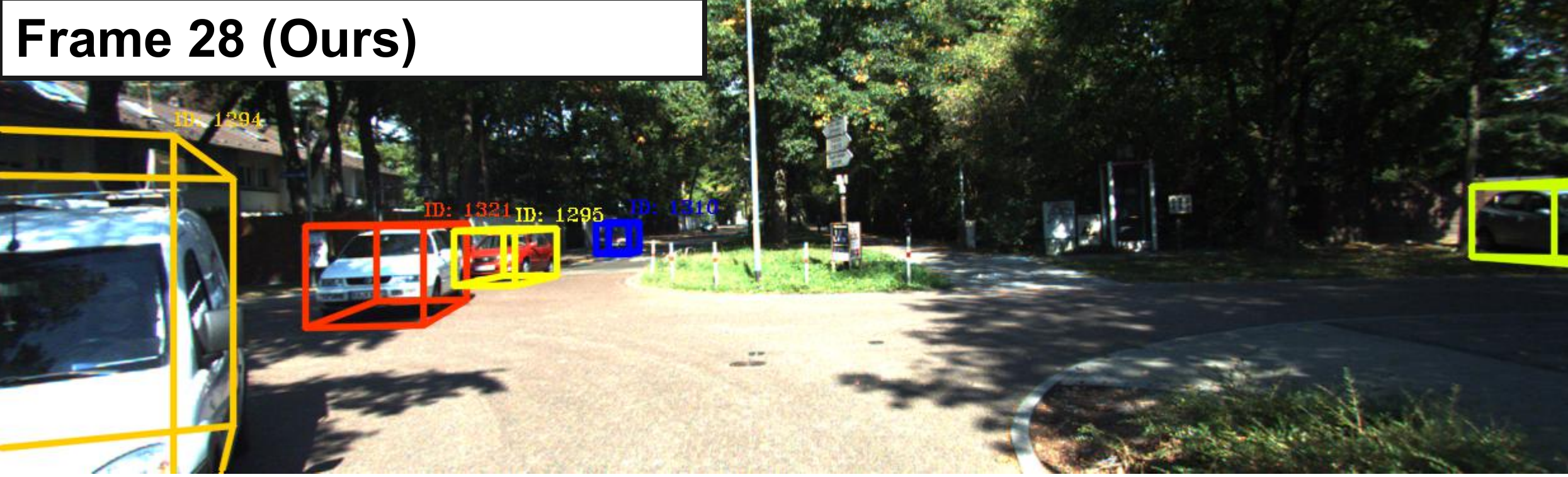}
\includegraphics[trim=0cm 6.6cm 0.1cm 0cm, clip=true, width=0.32\linewidth]{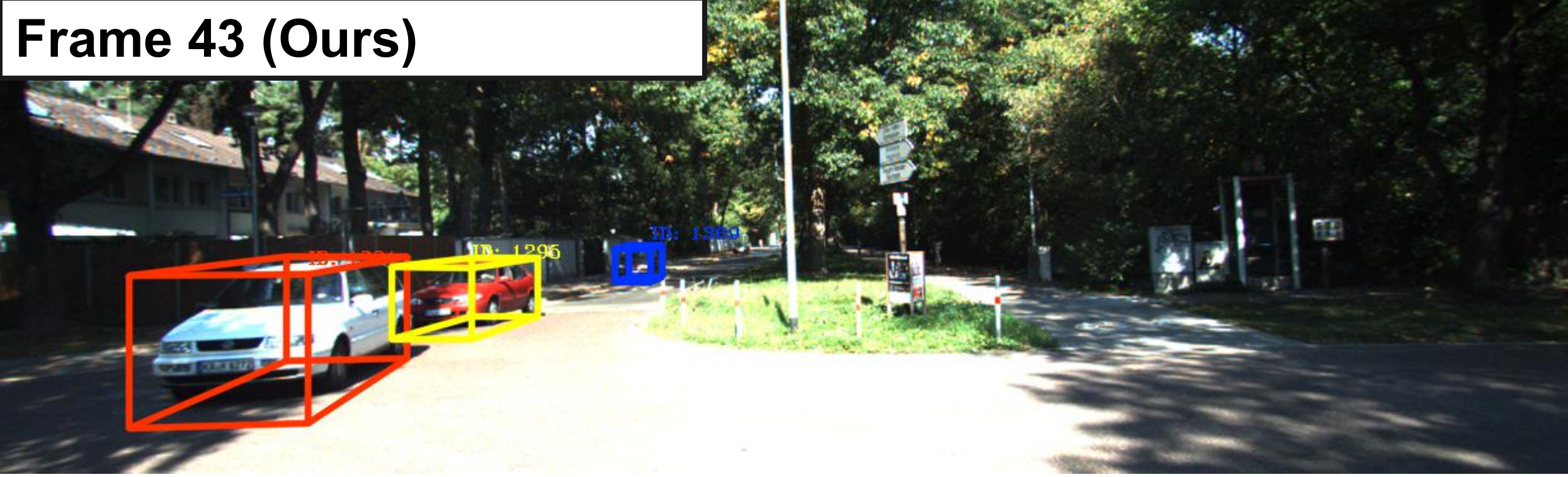}
\end{center}
\vspace{-0.75cm}
\caption{Qualitative results of our system on the sequence 3 of the KITTI \textbf{test} set.} 
\label{fig:qua}
\vspace{-0.5cm}
\end{figure*}

\noindent\textbf{Dataset and Evaluation.} We evaluate on the KITTI 3D MOT dataset, which provides LiDAR point cloud and ground truth 3D bounding box trajectories. As the KITTI test set only supports 2D MOT evaluation and its ground truth is not released to users, we have to use the KITTI val set for 3D MOT evaluation. Following prior work, we evaluate on the car subset of the KITTI dataset for comparison. In addition to the proposed three integral metrics, we also evaluate on the standard MOT metrics including MOTA, MOTP, IDS, FRAG, FPS.

\vspace{1.5mm}\noindent\textbf{Baselines.} We compare against recent open-sourced 3D MOT systems such as FANTrack \cite{Baser2019} and mmMOT \cite{Zhang2019}. We use the same 3D detections obtained by PointRCNN \cite{Shi2019} on KITTI for our proposed method and baselines \cite{Baser2019,Zhang2019} that require 3D detections as inputs. For baselines \cite{Baser2019,Zhang2019} that require the 2D detections as inputs, we use the 2D projection of 3D detections. 

\vspace{1.5mm}\noindent\textbf{Results}. We show the results of baselines and our proposed system in Table \ref{tab:3dcomparison}. Evaluation is conducted in 3D space using the proposed 3D MOT evaluation tool with new metrics. Our 3D MOT system consistently outperforms other modern 3D MOT systems in all metrics, establishing strong performance on KITTI 3D MOT and achieving an impressive zero identity switch. We show qualitative results of our 3D MOT system in Fig. \ref{fig:qua}. The 3D tracking results are visualized on the image with colored 3D bounding boxes where the color represents the object identity. We can see that our system can reliably track objects in the 3D space on the example sequence.

\vspace{1.5mm}\noindent\textbf{Inference Time.} We show comparison of inference time in last column of Table \ref{tab:3dcomparison}. Our 3D MOT baseline system (excluding the 3D detector part) runs at a rate of $207.4$ FPS on the KITTI val set without the need of GPU, achieving the fastest speed among modern 3D MOT systems.


\vspace{-0.4cm}
\section{Conclusion}
\vspace{-0.2cm}

We proposed an accurate, simple and real-time baseline system for online 3D MOT. Also, a new 3D MOT evaluation tool along with a set of new metrics was proposed for standardized 3D MOT evaluation in the future. Through experiments on KITTI 3D MOT benchmark, our system established strong 3D MOT performance while achieving the fastest speed. We hope that our system with released code will serve as a solid baseline on which others can easily build to advance the state-of-the-art in 3D MOT. Also, we hope that our released evaluation tool will serve as a standard in future 3D MOT benchmarks.

%
%

\vspace{-0.45cm}

\bibliographystyle{splncs04}
\bibliography{main}

\end{document}